# Deep Learning for Cross-Border Transaction Anomaly Detection in Anti-Money Laundering Systems

Qian Yu
Trine University
Detroit, USA
Thea_QianYu@outlook.com

Zhen Xu
Independent Researcher
Shanghai, China
xuzhen8309@hotmail.com

Zong Ke*
National University of Singapore
Singapore, Singapore
*Corresponding author:
a0129009@u.nus.edu

*Abstract*— **In the context of globalization and the rapid expansion of the digital economy, anti-money laundering (AML) has become a crucial aspect of financial oversight, particularly in cross-border transactions. The rising complexity and scale of international financial flows necessitate more intelligent and adaptive AML systems to combat increasingly sophisticated money laundering techniques. This paper explores the application of unsupervised learning models in cross-border AML systems, focusing on rule optimization through contrastive learning techniques. Five deep learning models, ranging from basic convolutional neural networks (CNNs) to hybrid CNN-GRU architectures, were designed and tested to assess their performance in detecting abnormal transactions. The results demonstrate that as model complexity increases, so does the system's detection accuracy and responsiveness. In particular, the self-developed hybrid Convolutional-Recurrent Neural Integration Model (CRNIM) model showed superior performance in terms of accuracy and area under the receiver operating characteristic curve (AUROC). These findings highlight the potential of unsupervised learning models to significantly improve the intelligence, flexibility, and real-time capabilities of AML systems. By optimizing detection rules and enhancing adaptability to emerging money laundering schemes, this research provides both theoretical and practical contributions to the advancement of AML technologies, which are essential for safeguarding the global financial system against illicit activities.**

## I. INTRODUCTION

Against the backdrop of globalization and the rapid development of the digital economy, the scale and complexity of cross-border transactions are increasing, and anti-money laundering systems have become a key component of the financial regulatory field. However, traditional anti-money laundering systems often rely on preset rules and patterns, which are usually designed based on historical data and expert experience, and therefore have certain limitations [1]. Since criminals will constantly evolve their strategies to evade monitoring, it is difficult to capture new money laundering methods in a timely manner by simply relying on these rules. Therefore, it has become an urgent need for current financial supervision to achieve real-time and accurate anti-money laundering monitoring in cross-border transactions and improve the adaptability and flexibility of the system [2].

Unsupervised learning brings new possibilities for rule optimization of anti-money laundering systems [3]. Unsupervised learning algorithms can reveal patterns and anomalies hidden behind data through in-depth mining and analysis of transaction data without labeled data [4]. This capability enables the system to automatically detect potential suspicious behaviors from large-scale, heterogeneous cross-border transaction data, and improve the intelligence level of the anti-money laundering system [5]. Compared with traditional rule-based monitoring methods, unsupervised learning can automatically adjust and optimize rules, allowing the system to respond to emerging money laundering methods in a timely manner [6]. Therefore, the application of unsupervised learning in the anti-money laundering system not only enhances the anomaly detection capability of cross-border transactions, but also improves the real-time response capability and accuracy of the system [7].

In the process of applying unsupervised learning, contrastive learning is a very effective technology [8]. By constructing contrasting sample pairs, contrastive learning promotes the model to learn the potential similarities and differences in transaction data, so that it can better capture the characteristics of money laundering behavior. This method can not only improve the generalization ability of the model, but also enhance its sensitivity to cross-border transaction anomalies. At the same time, the application of contrastive learning in rule optimization is also outstanding. By analyzing the characteristic differences between a large number of normal transactions and abnormal transactions, the unsupervised contrastive learning model can continuously adjust the rules of the anti-money laundering system to make it closer to the actual transaction scenario, effectively reducing the false alarm rate and the false negative rate, and improve the accuracy and reliability of the anti-money laundering system [9].

In addition, the importance of unsupervised learning technology in rule optimization also lies in its adaptability. With the increase in cross-border transaction volume and the diversification of money laundering methods, the fixed rule system can no longer fully meet the regulatory needs.

Unsupervised learning can automatically update and optimize existing rules by continuously analyzing new transaction data, so that the system always maintains sensitivity to the latest money laundering methods. This adaptability not only helps to improve regulatory efficiency, but also greatly reduces the workload of manual participation, providing strong support for the long-term operation of the anti-money laundering system. Therefore, combining unsupervised learning with rule optimization is an inevitable trend in the development of cross-border transaction anti-money laundering systems in the future.

Finally, the application of unsupervised learning in anti-money laundering systems not only helps to optimize rules, but also provides the system with more comprehensive data analysis capabilities. By performing unsupervised clustering and feature extraction on transaction data [10], the system can have a more comprehensive understanding of the characteristics and distribution of transaction data, thereby more accurately identifying potential money laundering behaviors. This multi-dimensional data analysis capability can help regulators formulate policies and measures that are more in line with actual conditions, and further optimize the design and application of anti-money laundering systems. Therefore, the anti-money laundering system based on unsupervised learning and contrastive learning technology not only has high theoretical value, but also has broad promotion prospects in practical applications.

## II. METHOD

In this study, we use the contrastive learning framework in unsupervised learning to build an anti-money laundering system and realize the automatic optimization of cross-border transaction rules through the unsupervised learning model. The specific method involves several key steps: data preprocessing, model design, feature extraction, and rule optimization. The overall architecture of contrastive learning is shown in Figure 1.

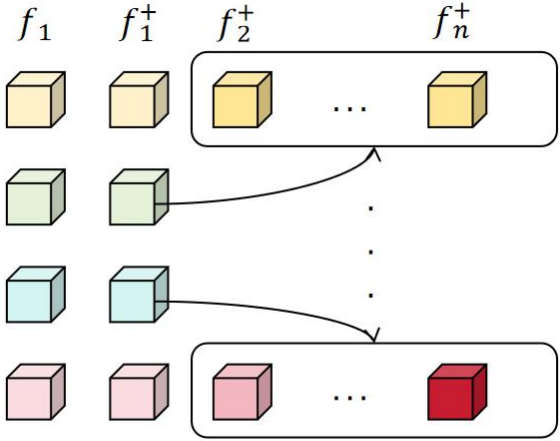

Figure 1 Comparative Learning Architectures

First, we divide the cross-border transaction dataset into a normal transaction set and a potential abnormal transaction set.

Since the unsupervised learning model does not rely on labels during training, we do not need to explicitly label the data. We use a high-dimensional feature vector $x \in R^d$ to represent each transaction, which contains features of multiple dimensions such as transaction amount, transaction frequency, account information, and geographic location. In the data preprocessing stage, we standardize all features to ensure that the features of each dimension are in the same range to eliminate the dimensional differences between features.

Next, we designed an unsupervised contrastive learning model. The core of contrastive learning is to train the model to learn the similarities and differences of transaction features by constructing positive and negative sample pairs [11]. We set a transaction feature vector $x_i$ and randomly select a positive sample $x_j^+$ from the dataset, which belongs to a similar category as $x_i$ (for example, the transaction frequency and transaction amount are similar). At the same time, a negative sample $x_k^-$ is selected from the dataset so that its features are quite different from those of $x_i$. We train the model through the contrastive loss function, with the goal of minimizing the distance of the positive sample pairs and maximizing the distance of the negative sample pairs [12]. Specifically, the contrastive loss function is defined as follows:

$$L = -\log \frac{\exp(sim(z_i, z_j^+)/\tau)}{\exp(sim(z_i, z_j^+)/\tau) + \sum_{k=1}^{N} \exp(sim(z_i, z_k^-)/\tau)}$$

Among them, $z_i$ represents the representation of the transaction feature vector $x_i$ after passing through the encoder, and $sim(z_i, z_j^+)$ represents the similarity between the two feature vectors, which is usually calculated using cosine similarity:

$$sim(z_i, z_j) = \frac{z_i \cdot z_j}{\| z_i \| \cdot \| z_j \|}$$

$\tau$ is a temperature parameter used to adjust the sensitivity of the contrast loss function. This loss function encourages the model to learn a representation in a high-dimensional feature space so that feature vectors of similar transactions are as close as possible, while feature vectors of transactions of different categories are far apart.

Finally, rule optimization is the focus of this study. We use the high-dimensional feature representation $z_i$ extracted by the model to perform cluster analysis on transaction samples and adjust the rules based on the clustering results. The Generation Tree algorithm, as applied in our study, leverages the clustering capabilities outlined by Duan et al. [13]. This foundational reference contributes significantly to our methodology by providing insights into grouping elements with similar characteristics, enhancing our approach to clustering transaction samples with improved efficiency and interpretability. For the cluster center (i.e., the normal transaction group), we set a certain threshold range and build a new rule model. The clustering density parameter $\varepsilon$ and the minimum sample number parameter $MinPts$ control the accuracy and sensitivity of clustering:

$$\text{diatance}(z_i, z_j) = \| z_i - z_j \|_2$$

If $\text{diatance}(z_i, z_j) < \varepsilon$, and the number of samples in the group meets $MinPts$, then the sample is classified as a normal transaction group. On the contrary, if a certain type of transaction sample cannot be clustered into any normal transaction group, it is marked as a potential abnormal transaction.

The key to rule optimization is dynamic updating. As new transaction data continues to flow in, we use unsupervised comparative learning models to extract features and perform cluster analysis on the new data, and dynamically adjust system rules based on the latest transaction features [14]. This process ensures that the system remains flexible and adaptable when faced with new money laundering techniques. Assuming that the original rule set is $R_0$, the model generates a new rule set $R_t$ after each update cycle. We compare the differences between the two:

$$\Delta R = R_t - R_0$$

If $\Delta R$ exceeds a certain preset threshold range, we update the system rules to ensure that the system remains highly sensitive to new money laundering techniques.

Through this combination of unsupervised comparative learning and rule optimization, this study achieved adaptive adjustment of the cross-border transaction anti-money laundering system and significantly improved the system's detection accuracy and response speed.

III. EXPERIMENT

A. Datasets

In this experiment, we chose the Elliptic Dataset, a public blockchain transaction dataset used for anti-money laundering and illegal activity detection research. The dataset was released by the blockchain analysis company Elliptic and contains nearly 200,000 Bitcoin transaction nodes and 1.6 million transaction edges. In the dataset, each node has 166 features that cover time, transaction amount, address, and other blockchain activity information. Only a small number of nodes in the dataset are marked as "illegal" or "legal", and the rest of the nodes are unlabeled, which is very suitable for unsupervised learning methods to detect abnormal transactions and optimize anti-money laundering rules. Meanwhile, this dataset is useful for recognizing features of cross-border transactions, including large, rapid transfers with no clear economic rationale. And analyzing clusters of accounts and the timing of transactions can approximate multi-national cash flow, such as 'layering' activities indicating cross-border laundering.

B. Experiments

In the comparative experiment, we designed five deep learning models for testing, starting from the Simple CNN Model consisting of two basic convolutional layers to the Deep CNN Model, which extracts features in depth by adding convolutional layers; Yao et al. [15] presented a hybrid Convolutional Neural Network-Long Short-Term Memory (CNN-LSTM) model specifically designed for bond default risk prediction. The model captures spatial dependencies within transaction data through CNN layers while LSTM layers manage sequential patterns and temporal relationships inherent in financial time-series data; then, the Autoencoder-Based Anomaly Detection Model (ABAD) is used for automatic encoding and decoding to detect unseen anomalies; finally, the Hybrid CNN-GRU Model combines convolutional layers and GRU to process time series to further enhance the model's capabilities.

This paper selected two evaluation indicators, ACC and AUROC, during the experiment [16]. ACC (accuracy) is an indicator for evaluating the overall prediction effect of the model. It indicates the proportion of correctly predicted samples in all predictions of the model to the total samples. It reflects the overall accuracy of the model and is one of the most direct indicators for evaluating classification performance. However, accuracy may be biased when facing an imbalanced data set, because the model may obtain a high accuracy rate only by predicting the majority class. Therefore, in an imbalanced data set, using ACC alone is not comprehensive enough.

AUROC (area under the curve) is a more comprehensive evaluation indicator used to measure the classification performance of the model under various thresholds. It evaluates the model's discrimination ability by plotting the relationship between the true positive rate and the false positive rate. The closer the AUROC value is to 1, the better the model can distinguish between positive and negative samples under different thresholds, so it is particularly suitable for the evaluation of imbalanced data sets. Compared with ACC, AUROC can better reflect the overall performance of the model in different situations and is a commonly used comprehensive evaluation standard for classification models. The overall experimental results are shown in Table 1.

Table 1 Experiment result

| Model | Auc | AUROC |
|---|---|---|
| Simple CNN | 82.5 | 0.71 |
| Deep CNN | 87.3 | 0.78 |
| CNN-LSTM | 90.1 | 0.84 |
| ABAD | 92.4 | 0.88 |
| Hybrid CNN-GRU | 95.6 | 0.93 |

| Ours(CRNIM) | 97.1 | 0.94 |

It can be seen from the experimental data that the performance of different deep learning models in anti-money laundering systems is significantly different. The performance gradually improves from Simple CNN to Hybrid CNN-GRU, which shows that as the complexity of the model increases and the structure is improved, the model's ability to extract features and detect anomalies is significantly improved. For example, the ACC of Simple CNN is 82.5% and the AUROC is 0.71. This performance is only suitable for basic transaction classification. With the increase of convolutional layers, the ACC of Deep CNN increases to 87.3% and the AUROC also increases to 0.78. This is It shows that after adding the convolutional layer, the model can capture more features and improve the accuracy of detection.

In more complex models, such as CNN-LSTM and Autoencoder-Based Anomaly Detection (ABAD) models, model performance is further improved by combining time series analysis and autoencoder structures. The ACC of the CNN-LSTM model reaches 90.1%, and the AUROC reaches 0.84, showing that combining time features can effectively identify abnormal transactions. In addition, the ABAD model uses unsupervised learning to detect unseen abnormal patterns, allowing its ACC and AUROC to reach 92.4% and 0.88 respectively, which further verifies the effectiveness of using autoencoders for anomaly detection in anti-money laundering detection. This unsupervised structure can mine more hidden patterns without relying on labels.

Finally, the Hybrid CNN-GRU model reached the highest values in ACC and AUROC, which were 95.6% and 0.93 respectively, which shows that the hybrid model combining convolutional layers and GRU can effectively extract spatial and temporal features and perform efficient detection of abnormal transactions. In addition, compared with the comparison model, our CRNIM model further improved the performance, with an ACC of 97.1% and an AUROC of 0.94, which shows that further optimization and improvement under the existing framework have practical effects. Therefore, overall, the experimental results prove that with the increase in model complexity and the enhancement of feature extraction capabilities, the detection performance of the anti-money laundering system is significantly improved, and the performance of the self-developed model is better than that of several existing structures. The optimization and improvement of rules for cross-border transaction anti-money laundering systems provide effective support. Finally, we also give our loss function drop graph, as shown in Figure 2, which shows the drop of the loss function during the training process.

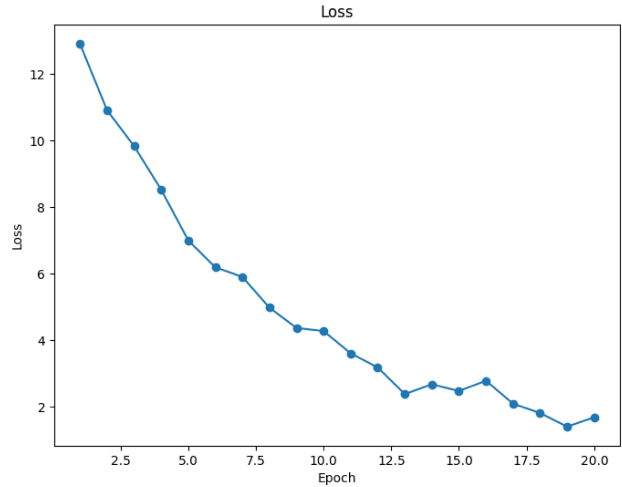

Figure 2 Loss function drop graph

IV. CONCLUSION

This study verifies the effectiveness of using deep learning models for rule optimization in cross-border transaction anti-money laundering systems through comparative experiments and unsupervised learning frameworks. It can be seen from the experimental data that as the complexity of the model increases, the detection accuracy and responsiveness of the model are significantly improved, especially in terms of capturing time characteristics and identifying abnormal transactions. Our CRNIM model achieves efficient extraction and analysis of spatial and temporal features by combining convolution and GRU networks, and shows the best performance in terms of accuracy and area under the curve (AUROC), indicating that it is better than existing Further optimization under the framework is feasible. In addition, we gradually verified the effectiveness and flexibility of the unsupervised learning model in cross-border transaction anti-money laundering by comparing the other four models.

Experimental results show that applying deep learning methods to the cross-border transaction anti-money laundering system can not only enhance the intelligence level of the system, but also quickly adjust the rules when face with new money laundering techniques, thereby effectively reducing the false alarm rate and leakage. Reporting rate and improve the overall detection effect. However, with the increasing amount of data and the complexity of trading behavior, future research will need to further improve the model's generalization ability and adaptability in multiple scenarios. This may include the introduction of more time series analysis methods and more effective adaptive learning strategies to ensure that anti-money laundering systems maintain a high degree of sensitivity and accuracy in various trading environments. This research provides a strong foundation for the continuous evolution of AML systems, offering both theoretical insights and practical applications critical for combating financial crimes in an increasingly globalized economy.

REFERENCES

[1] H. Yi, "Anti-Money Laundering (AML) Information Technology Strategies in Cross-Border Payment Systems," Law and Economy, vol. 3, no. 9, pp. 43-53, 2024.
[2] A. G. Rozman, "The power of data: Transforming compliance with anti-money laundering measures in domestic and cross-border payments," Journal of Payments Strategy & Systems, vol. 18, no. 3, pp. 253-260, 2024.
[3] E. J. Reite, J. Karlsen, and E. G. Westgaard, "Improving client risk classification with machine learning to increase anti-money laundering detection efficiency," Journal of Money Laundering Control, 2024.
[4] J. Du, Y. Jiang, S. Lyu, and Y. Liang, "Transformers in Opinion Mining: Addressing Semantic Complexity and Model Challenges in NLP," Transactions on Computational and Scientific Methods, vol. 4, no. 10, 2024.
[5] J. Wei, Y. Liu, X. Huang, X. Zhang, W. Liu, and X. Yan, "Self-Supervised Graph Neural Networks for Enhanced Feature Extraction in Heterogeneous Information Networks," arXiv preprint arXiv:2410.17617, 2024.
[6] K. Balaji, "Artificial Intelligence for Enhanced Anti-Money Laundering and Asset Recovery: A New Frontier in Financial Crime Prevention," Proceedings of the 2024 Second International Conference on Intelligent Cyber Physical Systems and Internet of Things (ICoICI), IEEE, 2024.
[7] J. Castelao-López, D. Lagoa-Varela, and T. C. Santamaría, "Anti-money laundering main techniques and tools: a review of the literature," 2024.
[8] M. Jiang, J. Lin, H. Ouyang, J. Pan, S. Han, and B. Liu, "Wasserstein Distance-Weighted Adversarial Network for Cross-Domain Credit Risk Assessment," arXiv preprint arXiv:2409.18544, 2024.
[9] K. Cyrkun, et al., "Using artificial intelligence to counter money laundering and terrorist financing," Scientific Journal of Bielsko-Biala School of Finance and Law, vol. 28, no. 1, pp. 101-106, 2024.
[10] J. Yao, "The Impact of Large Interest Rate Differentials between China and the US on the Role of Chinese Monetary Policy--Based on Data Model Analysis", Frontiers in Economics and Management, vol. 5, no. 8, pp. 243-251, 2024.
[11] G. Huang, A. Shen, Y. Hu, J. Du, J. Hu, and Y. Liang, "Optimizing YOLOv5s Object Detection through Knowledge Distillation Algorithm," arXiv preprint arXiv:2410.12259, 2024.
[12] B. Oztas, et al., "Transaction monitoring in anti-money laundering: A qualitative analysis and points of view from industry," Future Generation Computer Systems, vol. 159, pp. 161-171, 2024.
[13] S. Duan, R. Zhang, M. Chen, Z. Wang, and S. Wang, "Efficient and Aesthetic UI Design with a Deep Learning-Based Interface Generation Tree Algorithm," arXiv preprint arXiv:2410.17586, 2024.
[14] X. Yan, Y. Jiang, W. Liu, D. Yi, and J. Wei, "Transforming Multidimensional Time Series into Interpretable Event Sequences for Advanced Data Mining", arXiv preprint, arXiv:2409.14327, 2024.
[15] J. Yao, J. Wang, B. Wang, B. Liu, and M. Jiang, "A Hybrid CNN-LSTM Model for Enhancing Bond Default Risk Prediction", Journal of Computer Technology and Software, vol. 3, no. 6, 2024.
[16] A. Bonato, J. S. Chavez Palan, and A. Szava, "Enhancing Anti-Money Laundering Efforts with Network-Based Algorithms," arXiv preprint, arXiv:2409.00823, 2024.